# Plan Recognition in Stories and in Life*

Eugene Charniak (ec@cs.brown.edu) and Robert Goldman (rpg@cs.brown.edu)
Department of Computer Science, Brown University
Box 1910, Providence RI 02912
(401) 863-7600

## Abstract

Plan recognition does not work the same way in stories and in "real life" (people tend to jump to conclusions more in stories). We present a theory of this, for the particular case of how objects in stories (or in life) influence plan recognition decisions. We provide a Bayesian network formalization of a simple first-order theory of plans, and show how a particular network parameter seems to govern the difference between "life-like" and "story-like" response. We then show why this parameter would be influenced (in the desired way) by a model of speaker (or author) topic selection which assumes that facts in stories are typically "relevant".

## 1 Introduction

Plan recognition in stories does not work in the way it does in "real life." Consider:

> Jack wanted to kill himself. A pistol dating back to the days of Teddy Roosevelt and the Rough Riders was a family heirloom.

If you encountered this in a story you would seriously entertain the idea that Jack might try to shoot himself. If asked to quantify your belief you would say that the chances were less than one, but more than, say, .2. You would presumably say the same here:

> Jack wanted to kill himself. There was a rope in the closet.

Now, of course, Jack would be hanging himself.

Life works differently. If the facts in the first story were true of a friend of yours you would not really believe that he would use the gun. After all, he probably has rope, knives, pills, and a high building to jump from. (This is not to say that shooting might not occur to you, but simply that you would not believe that he would do it with anything like a 20% probability.) We assume that stories give us selected, typically relevant, information, while life gives us unselected, mostly tangential, facts. Somehow this affects plan recognition in the two areas.

This is not news to those of us working in plan recognition in stories. However as it complicates things, and makes what we are doing seem more ephemeral, we steadfastly ignore the fact. Indeed, the only published reference to the problem we have found is (Kaplan [1978]). Kaplan proposes a class of "language driven inferences" which are distinct from "domain driven inferences." However, since he does not specify what these are, this is really a restatement of the problem. We all agree that we reach extra conclusions in stories, and thus must be making extra inferences. The question is *why*? (Not to mention *which ones*, and *how*?)

We take the position that plan recognition is done the same way (in people) for both stories and life. Therefore we are looking for a single theory which can handle both, and explains why and how they differ. This paper will present such a theory, at least for the case of how objects in a story influence our perceptions of people's plans in a way different from the object's presence in the world. In Section 2 we formulate the connection between plans and their objects in a simple first-order version of frames and slots. Section 3 then recasts this in probabilistic terms, using Bayes networks. We will see how by setting a particular network parameter (the prior probability that two things are the same) to different values we can get both story-like and life-like interpretations out of the network, and thus in a crude way shows how a single system could handle both cases. Section 4 improves this theory by showing how reasonable assumptions about speaker's intentions, "only mention relevant things," will also modify this parameter. We thereby get the theory we want: one which handles both stories and "life", and explains the difference in terms of a simple assumption about selection in the case of stories.

*This work has been supported in part by the National Science Foundation under grants IST 8416034 and IST 8515005 and Office of Naval Research under grant N00014-79-C-0529.

## 2 Preliminaries

### 2.1 Objects and Plans in a First-order Frame Theory

We adopt a simple model of plan recognition in which plans are complicated actions (ones with substeps) and all plans are built out of a complete library of plan schemas. Plan schemas are natural kinds ("frames") and the steps of a plan are "slots" in frames. In the first order reconstruction, these slots are functions from instances of a plan schema to the actions which accomplish the step in the plan. To take an example we will be using a lot, here is how we would say that some person j1 is engaged in a hanging plan h2, and that a particular getting event g3 filled the step in the plan by which the agent gets the necessary rope (which we will call the get-step):

(person j1)
(hang h2)
(= (agent h2) j1)
(get g3)
(= (get-step h2) g3)

The last line here serves as the explanation of the getting event in terms of a higher plan. Objects in plans are also connected to the plan via slots (functions). So the rope-of slot would associate a hanging plan with the rope used in it, as in (= (rope-of h2) r4). We assume that all slots have restrictions on what can fill them; rope-of must be filled by a rope. This is expressed as the general constraint (hang ?x)→(rope (rope-of ?x)). Throughout the rest of this paper we will be considering the plan recognition problems inherent in the two stories:

"There was a rope in the closet. Jack killed himself."

"Jack got a rope. He killed himself."

We take it for granted that the second example differs from the first in that in real life we would assume that Jack hung himself.

It is important to note that nothing we say is particular to these examples. So, more generally, note that if an object $j$ fills the slot $s$ in the instance $i$ of the plan schema $p$, where $s$ must be filled with entities of type $t$, the following will be true: $(p\ i)$, $(=(s\ i)\ j)$, and $(t\ (s\ i))$. These are the only facts we will be using in our discussion of ropes and hangings.

### 2.2 Bayesian Networks

Bayesian networks are a way of representing information about probability distributions, and are particularly convenient for representing networks of causal influences.[1] The idea is that some set of random variables (for us, boolean variables, denoting partial states of affairs, existence of actions, objects, etc.) influence other random variables. We write the variables as nodes, and the influences as as directed arcs, with the direction indicating causality (or picked for convenience in non-causal situations). Figure 1 shows a piece of a medical diagnosis network. (In "right heart syndrome" the heart has trouble pumping blood, and some backs up into the lungs.)

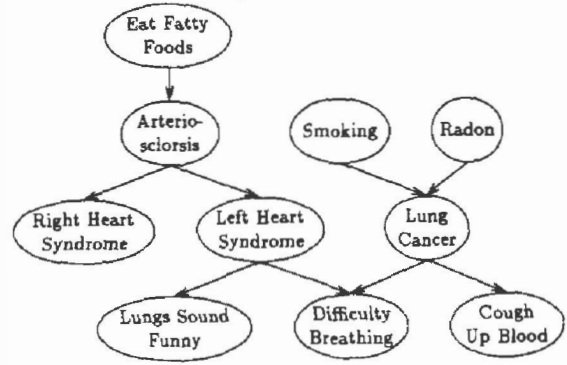

Figure 1: Bayesian network for medical diagnosis

We will not attempt to defend our use of probability theory in plan recognition other than to note that like all abductive tasks, plan recognition requires putting together many pieces of evidence into a probable, but not certain, conclusion. Probabilities are an obvious candidate for such a task.

We use Bayesian networks to represent our probability distributions because of the way they summarize dependencies and independencies. We chose them over Markov networks, the undirected graph representation, because it is easier to specify the probabilty distribution corresponding to a directed graph. Influence diagrams are generalizations of Bayesian networks which include nodes for decisions and utilities.[2] We do not need this much expressive power because our program is simply a recognizer: it neither acts in the world, nor does it model the decision-making process of agents in the world.

It is beyond the scope of this paper to discuss the tractability of computing probability distributions in these networks. In general the problem is intractable. (Cooper [1987]) If one has a singly connected network (at most one undirected path between any two nodes) there is a polynomial algorithm due to Pearl and Kim (Pearl [1988]). We expect that the networks we need will, in general, be multiply connected, and we are looking for ways to approximate the probability distribution.

---

[1] A thorough introduction to Bayesian networks may be found in (Pearl [1988]).

[2] See, e.g., (Schachter [1986]) for an introduction to influence diagrams.

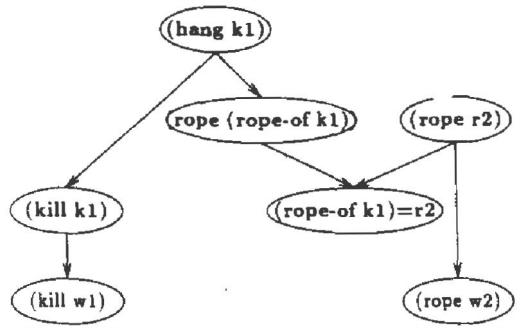

Figure 2: Partial network for the "rope in the closet" example

Finally, we would like to make it clear that while this is a probabilistic approach, it is not a statistical one. Rather, our approach is axiomatic and logical. While we would like to be able to collect statistics for the probabilities we need, we have no idea how to go about doing this.

## 3 The "Knob" Theory

Now we combine our first-order formalism for plans and their associated objects with Bayesian networks. Figure 2 gives our Bayes network reconstruction of the "rope in the closet" example, though to keep things simple it ignores Jack's role, just concentrating on the clues provided by the words "kill" and "rope." The link between (kill k1) and "kill" represents the idea that the presence of a killing (in the story) caused the use of the word "kill."[3] Figure 2 also represents the connection between the hanging event and the killing, and the connection between the presence of a rope and the word "rope." The most interesting part, however, is the connection between the hanging event, and the rope. As in our first-order theory Figure 2 says that the presence of the hanging implies the presence of *some* rope, which is denoted by the term (rope-of k1). However, this might or might not be the same as the one in the story, r2. Thus, the presence of a hanging event will only predict the presence of r2 if the two are the same, (*i.e.*, (= (rope-of k1) r2)). Thus both of these condition our expectation that r2 is a rope. In particular, if both hold then (rope r2) holds with probability 1.

---

[3]Some people have trouble with the causal interpretation of the links between states and events in the world, and the words in the text, as when we said that the killing event caused the use of the work "kill." The way to think about this is that we have an observer who is writing down what he or she observed. Thus the observation of a killing caused her to write it down, and thus use this word. Obviously the real story is much more complex since it is impossible to write down everything one observes. Thus our causal connection is probabilistic, reflecting the fact that we do not have all of the information and causal influences specified (and probably never could). The same is true in the medical diagnosis case.

To complete Figure 2 we also need probabilities (the prior probabilities on the topmost nodes, and conditional probabilities on child nodes given the parent nodes). The particular probability which will prove most important for our discussion is the prior for (= (rope-of k1) r2). Before we can say much about it, however, we must figure out what such a probabilistic node *means*. In (Charniak and Goldman [1989]) we discuss this issue. The key point here is that terms like k1 and r2 are to be interpreted as the outcomes of experiments, where the sample space is all the things in our domain: objects, actions, whatever. In other words, these are arbitrary symbols and could, in principle, denote anything. Of course, since we postulate that they are the referents of words in our text, and since the words only can refer to certain kinds of things (the word "rope" cannot refer to a tomato, or at least the probability of this is *very* low) the words used start the process of pinning down exactly what the terms denote. Similarly, n-ary function symbols (*e.g.*, rope-of) denote functions from n members of the sample space to a member of the sample space. The point here is that until we know more, the probability of (= (rope-of k1) r2) is *very* low. It is the probability that two arbitrary entities will turn out to be the same. If you imagined a world populated with, say $10^{20}$ entities, then this would be on the order of $10^{-20}$. Also note that any equality statement will have this same prior probability, and thus we are not talking about something particular to this example, but rather a "fundamental constant" of the system, henceforth denoted $E$. It says, in effect, how likely things are to be associated. (Remember, in our first-order frame representation, associating one entity with another is having one fill a slot in another, and this is expressed using equality).

(Charniak and Goldman [1989]) also discusses the value of $E$, and points out that in realistic stories and realistic observations there is temporal and spatial locality, in the sense that objects observed in sequence will come (typically) from the same portion of "space-time". This raises the probability that two things will be equal, because when restricted to a small part of space-time there are fewer different objects around. In (Charniak and Goldman [1989]) we claim that when this constraint is taken into account, networks for our prototypical stories give reasonable "life-like" probabilities. That is, in the first the probability of hang is low, in the second it is high.[4] For this paper the reader will have to take this on trust. In what follows we will assume that the spatial/temporal locality constraint has already been taken into account.

We discuss this particular "fundamental constant" ($E$) because changing it will cause the network to assign values to the probability of (hang k1) which range from the story-like, on one hand, to life-like on the other. To see why this is the case it is necessary to have some feel for the flow of probabilities in the network. Setting "rope" and "kill" to 1 causes (rope r2) and (kill k1) to now have probability 1. The latter will push the probability of (hang k1) to a number like, say, $10^{-3}$. This would be saying that one out of a thousand killings is a hanging. The question then becomes, will the evidence from (rope r2) cause it to go even higher. If $E$ is very low (that is, lower than the prior probability of something being a rope) the answer is no. In essence given this low value, it is more likely that a rope appeared in the story "by accident" than it's presence be depending on this equality statement. But as we raise the value of $E$ it becomes easier and easier to explain (rope r2) on the basis of assuming both (rope (rope-of k1)) and (= (rope-of k1) r2). This then raises the probability of (rope (rope-of k1)), and in turn (hang k1). More formally, let $e$ denote (= (rope-of k1) r2), $k$ denote (rope (rope-of k1)), and $r$ denote (rope r2). Then $P(r \mid e, k) = 1$, $P(r \mid e, \neg k) = 0$, and $P(r \mid \neg e, k) \approx P(r \mid \neg e, \neg k) = P(r)$ If we assume, as will be the case here, that $P(r) \ll 1$ and $P(k) \ll 1$, then at the point where $P(e) = P(r)$, $\frac{P(k|r)}{P(k)} = 2$. (The proof just takes $P(k \mid r)$, expands it using Bayes theorem, and expands out terms involving $r$ using the above conditional probabilities along with the independence assumptions implicit in Figure 2.) In other words, by setting $E$ to be the prior probability that something is a rope we will double the posterior probability that (rope (rope-of k1)). This will have a similar effect on the probability of hanging. Increasing $E$ still further will have an approximately linear effect on the probability of hanging until the later starts to approach one. Thus at high values of $E$ we get story like interpretations, while when $E$ at or below $P(r)$ we get "life-like" interpretations.[5]

This then gives us our first, crude, theory for combining both "story-like" and "life-like" plan recognition in a single system. By adopting the Bayesian framework given here the parameter $E$ becomes a "knob" which can be adjusted to give the kind of system we want, a story reader or a real-life plan recognizer.

---
[4]Those who do consult (Charniak and Goldman [1989]) will note that the networks there differ from the ones here in the placement of the equality node. Nothing in either theory depends on this difference. Rather differing pedagogical needs dictated the difference.

[5]Note that turning $E$ beyond $P(r)$ will still not necessarily be sufficient to hook up very common things, like, say, leaves, because their prior probability is still higher. This seems reasonable.

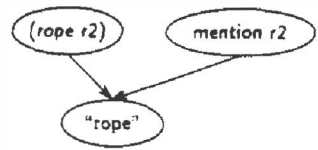

Figure 3: Why one uses a particular word.

## 4 The "Mention" Theory

The "knob" theory shows that it is possible to have a unified theory, but it does not explain why it is that we needed two theories at all. Why is it that plan recognition differs in stories and life? The knob theory takes it for granted that we need two versions of plan recognition and just shows how to model them in a single system.

We do have some intuitions about why plan recognition is different in stories. As we said in the introduction, in stories things are expected to be relevant, in life they are not. Or to put this another way, we have beliefs about what authors chose to include in a text and what they leave out. They include the relevant, interesting things, and delete the rest. This suggests that we progress toward an explanatory theory by including more of the factors which lead to things being put into the text.

We start by looking more closely at why a writer uses a certain word (e.g., "bank") in a text. In our very simplified model the writer uses a word because the object in question, first-national-43, is of a particular type, savings-institution, and a word of English, "bank," is used to denote such objects. While this causal connection is necessary, it is not sufficient. In our simplifications we have chosen to ignore another necessary component behind the use of the word "bank," the idea that the writer wanted to mention first-national-43. Let us now complicate our lower portions of the Bayesian network to look like the fragment in Figure 3. Needless to say, as a theory of word-choice, sentence generation, etc., this is still pretty poor, but it is enough for our current purposes.

Next we want to ask why someone will mention a particular entity. This is a complicated matter, and we do not have a theory. Rather we want to show that there is good reason to believe that if we had such a theory it would take over "turning the knob." That is, there would be connections in our Bayesian network between the node (= (rope-of k1) r2) and mention r2. To see this, consider the quite unobjectionable rule, "objects used in already mentioned plans are more worthy of mention that other objects (everything else being equal)." This is not much of a theory of "mention," but it is hardly objectionable. Figure 4 shows the "rope in the closet" network with this rule added in. Note that this rule connects the equality statement with "mention." The reason is, of course, that in our theory we state that an object is used in a plan by asserting an equality between mentioned objects and the plan-role. That is ($=$ (rope-of k1) r2) states that r2 fills the rope-of slot in some plan. Now suppose that objects filling roles in already mentioned plans are $k$ times more likely to be mentioned than an arbitrary object. That is

$$\frac{P(mention | role\ in\ plan)}{P(mention)} = k$$

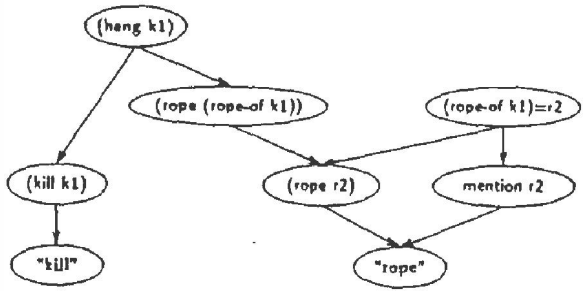

Figure 4: Mentioning the rope in the closet

An application of Bayes formula is sufficient to establish

$$\frac{P(role\ in\ plan | mention)}{P(role\ in\ plan)} = k$$

That is to say, the probability of the equality statement goes up by $k$ given that the word has been mentioned. Thus the fact that the author has mentioned an object is grounds for increasing the (posterior) probabilities of the equality statements. Furthermore, our guess is that $k$ is a very large number, since very few objects in a scene will be mentioned, while a good fraction of those involved in plans will be mentioned. Thus, even an unsophisticated rule like this may well be sufficient to modify the probability of the equality to give us "story like" performance for the class of inferences discussed here. Whether or not this particular "mention" rule is sufficient, this suggests that some such rule will be.

## 5 Conclusion

It seems clear that we can expand this theory a great deal without much change. So, while we have discussed inferences about the presence of objects, the theory should work for some inferences about states and events as well. Consider a story like:

Jack took Mary to a restaurant. Then menu was in French.

Lots of people go to fancy restaurants and most of the time they have no trouble reading the menu. However in this story we wonder if there will be a problem. Our rule for objects filling plans is clearly inadequate for this example, but one which discussed causal consequences of states would be reasonable Thus we have the beginnings of a theory for why inference in text comprehension is different from real life. It assumes there is a single parameter which governs the difference (for the general case there could well be a small set of parameters) and that assumptions about why authors include facts in texts modify the posterior probability of the parameter(s) in stories. For those (like the authors), who believe that plan recognition in life must be evolutionarily prior with the story process parasitic off of it, the fact that this theory takes the real-life case as basic is also satisfying.

## Appendix: Work in Progress

Currently, we are testing the theory discussed here by implementing a story understanding program based on it. Our program parses simple English stories like those discussed in this paper. Using network-construction rules akin to the forward-chaining rules of a data-dependency network, it incrementally builds Bayesian networks corresponding to the story (the "real-life" version at the moment). These networks are evaluated using an implementation of the algorithm (Lauritzen & Spiegelhalter [1988]) provided by Srinivas & Breese [1989]. The program is working, but its network evaluation is still too slow to be practical. We are investigating other ways of evaluating our belief networks.